\documentclass[10pt,twocolumn,letterpaper]{article}

\usepackage{iccv}
\usepackage{times}
\usepackage{epsfig}
\usepackage{graphicx}
\usepackage{amsmath}
\usepackage{amssymb}

\usepackage{balance}

\usepackage{caption}
\usepackage{color}

\definecolor{red}{rgb}{1.00,0.00,0.00}
\definecolor{blue}{rgb}{0.00,0.00,1.00}
\definecolor{green}{rgb}{0.00,1.00,0.00}

\newcommand{\cblue}[1] {\textcolor{blue}{#1}}

\usepackage[pagebackref=true,breaklinks=true,letterpaper=true,colorlinks,bookmarks=false]{hyperref}

\iccvfinalcopy 

\def\iccvPaperID{33} 

\ificcvfinal\fi

\begin{document}

\title{Tiny and Efficient Model for the Edge Detection Generalization}


\author{Xavier Soria\\
National University of Chimborazo, Ecuador\\
{\tt\small xavier.soria@unach.edu.ec}
\and
Yachuan Li\\
\normalsize China University of Petroleum (East China), China \\
{\tt\small liyachuan@s.upc.edu.cn}
\and
Mohammad Rouhani \\
INRIA Paris, France\\
{\tt\small mohammad.rouhani@inria.fr}
\and
Angel D. Sappa$^{1,2}$ \\
$^{1}$ESPOL Polytechnic University, Ecuador\\
$^{2}$Computer Vision Center, Spain\\
{\tt\small asappa@espol.edu.ec \& asappa@cvc.uab.es}
}

\maketitle
\ificcvfinal\thispagestyle{empty}\fi

\begin{abstract}

Most high-level computer vision tasks rely on low-level image operations as their initial processes. Operations such as edge detection, image enhancement, and super-resolution, provide the foundations for higher level image analysis. In this work we address the edge detection considering three main objectives: simplicity, efficiency, and generalization since current state-of-the-art (SOTA) edge detection models are increased in complexity for better accuracy. To achieve this, we present Tiny and Efficient Edge Detector (TEED), a light convolutional neural network with only $58K$ parameters, less than $0.2$\% of the state-of-the-art models. Training on the BIPED dataset takes \textbf{less than 30 minutes}, with each epoch requiring \textbf{less than 5 minutes}. Our proposed model is easy to train and it quickly converges within very first few epochs, while the predicted edge-maps are crisp and of high quality. Additionally, we propose a new dataset to test the generalization of edge detection, which comprises samples from popular images used in edge detection and image segmentation. The source code is available in \url{https://github.com/xavysp/TEED}.
 
\end{abstract}

\section{Introduction}
\label{sec:intro}

\begin{figure*}[t]
  \centering

   \includegraphics[width=\textwidth]{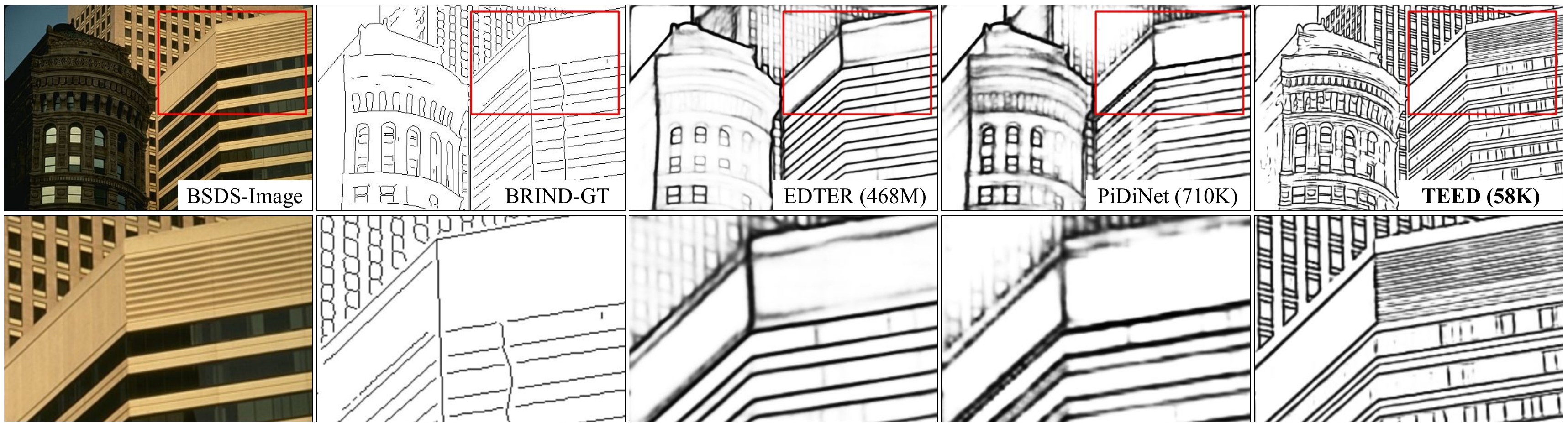}

   \caption{Edges from our proposal (TEED) and the state-of-the-art models. EDTER \cite{Pu2022edter} and PiDiNet \cite{su2021pidinet} have been trained in BSDS500\cite{arbelaez2011bsds500} following the standard training procedure. TEED has been trained \textbf{from the scratch with BIPED} \cite{soria2023dexined_ext} with a reduced hyper-parameters tuning. }
   \label{fig:banner}
\end{figure*}

Large scale Deep Learning (DL) models are frequently used in many computer vision applications, as documented in \cite{zhai2022scalingVT,khan2022ViTsurvey}. However, for low level tasks such as image enhancement, super-resolution \cite{yang2019SRreview} and edge detection \cite{yang2022edgeOverview}, more efficient and lightweight models are necessary as these steps are preliminary to higher level image analysis. Therefore, edge detection models should come with low computational cost and latency. For these reasons, classical edge detectors like Sobel \cite{sobel1972sobelmethod} or Canny \cite{canny1986canny} are still widely used in many applications. However, recent deep learning architectures with over 10 million parameters have been proposed to outperform state-of-the-art approaches in various benchmarks \cite{xie2015hed, Pu2022edter}. While these models are powerful, they come with a significant computational expense.



In order to reduce the computational cost, new procedures for training the DL models have emerged, allowing the utilization of lightweight models through careful dataset selection. According to \cite{Hou_2013NObsds, li2019noBSDS, soria2020dexined}, the standard datasets for edge detection, like BSDS \cite{arbelaez2011bsds500}, are originally introduced for image segmentation; although they have edge level annotations some of their ground truth comes with wrong annotations \cite{Hou_2013NObsds, soria2023dexined_ext}. Having this problem in mind, the new DL training procedure uses BIPED dataset instead of BSDS, which avoids tedious setting for transfer learning, and also reduces the training and testing time; we refer to this procedure as Training from the Scratch (TFS). The edge-maps generated from these different training procedures are compared in Fig. \ref{fig:banner}, where the results from TEED use the TFS procedure. Note that BSDS dataset was not used in training of TEED but our model is capable of generalize the edge detection. In this manuscript the edge detection generalization is the capacity of the learning algorithm to predict most edges from an arbitrary image---any image even gray-scale that comes from a visible wavelength band.


The proposal, named Tiny and Efficient Edge Detector (TEED) is capable of predicting thinner and clearer edge-maps. Compared to the SOTA models \cite{he2022bdcn-ext}, \cite{su2021pidinet}, and \cite{Pu2022edter} TEED stands out due to being remarkably \textbf{simple}, eliminating the need for transfer learning or exhaustive hyper-parameter tuning. Additionally, TEED is highly \textbf{efficient}, demonstrating rapid convergence and producing superior results both quantitatively and qualitatively. Our model generates robust results while it can handle various scenes and different input types, whether in color or grayscale. To assess the TEED's \textbf{generalization} ability, we have prepared a new test dataset by selecting images based on a \textit{frequency criteria} from commonly used datasets in edge detection, segmentation, and other low-level processing tasks. We named this collection of images the Unified Dataset for Edge Detection (UDED).


Overall, we present five main contributions: ($i$) TEED: a simple but robust CNN model with only 58K parameters; ($ii$) dfuse: a new, efficient fusion module inspired from CoFusion in CATS \cite{huan2022cats} (while coFusion has around 40K parameters, dfuse has less than 1K parameters); ($iii$) a dloss function for efficient and rapid training convergence; ($iv$) UDED: a new dataset to test edge detection generalization, which includes the ground truths annotated through human perceptual edges; and finally, ($v$) a fair quantitative and qualitative comparison with SOTA models that have less than 1M parameters in UDED. To evaluate the robustness in UDED dataset, a downstream task (sketch image retrieval) is used, this validation also compare the standard and new DL based approaches.


The remainder of this paper is organized as follows: Section \ref{sec:rw} reviews related work and discusses the parameter requirements of each approach. Section \ref{sec:teed} elaborates on the proposed architecture. Section \ref{sec:data} describes the dataset and the evaluation procedure. Section \ref{sec:expe} presents  experimental results; and finally, conclusions are given in Section \ref{sec:conc}.




\section{Literature Review}\label{sec:rw}

Edge detection is widely used from the low to high level image analysis (e.g., medical image segmentation \cite{du2022EdgeMedSeg}, sketch-based image retrieval \cite{song2017edgeSketch2}). For a comprehensive review, we refer readers to \cite{ziou1998edgeOverview1, Basu2002edge_survey, maini2009edgeOverview3}, and \cite{yang2022edgeOverview}.


In recent years different deep learning based approaches have been proposed for tackling the edge detection problem. They can be classified into two categories. The first category includes approaches that train their models mainly with BSDS500 \cite{arbelaez2011bsds500}, NYUD \cite{Gupta_2013NYUDv2}, and PASCAL-Context \cite{mottaghi2014PASCALcontext} databases, without applying a validation process on the given annotations. This leads to tedious additional steps before and during training. The second category includes approaches that split the problem up into edge, contour, and boundary detection, as also outlined in BSDS300 \cite{martin2004mBSDS300ext} and thoroughly explained in \cite{Hou_2013NObsds} and \cite{mely2016mdbd}. Subsequently, BIPED \cite{soria2020dexined} and BRIND \cite{pu2021brind} showed that a model trained from scratch on a curated dataset for edge detection could accurately predict over $80$\% of edges in a given scene. Our approach is aligned with this latter methodology. Additionally, we suggest that an efficient yet lightweight model, after training, can predict over $80$\% of edges in any image dataset considered for evaluation.


With recent developments in datasets proposed for DL model training, such as those in \cite{soria2023dexined_ext, pu2021brind}, the next step is to find a dataset that encompasses images from various scenarios for comprehensive evaluation. To this end, our paper introduces a small yet diverse dataset for edge evaluation. In addition, various metrics are considered to ensure a fair comparison, as presented in Section \ref{sec:expe}.


\subsection{Edge Fusion Methods}

Fusion of multi-scale features to generate edges is as important as feature extraction in edge detection tasks. The earliest methods involve reshaping the multi-scale feature matrix into the size of the final result, and then computing the weighted sum as the final result. This fusion method, though simple, is very effective and widely adopted by different works such as HED \cite{xie2015hed}, RCF \cite{liu2019RCFext}, BDCN \cite{he2022bdcn-ext}. However, this fusion approach has two drawbacks: firstly, the fine branches lack global semantic information, and secondly, features in the same channel share the same weight and have equal importance in channel fusion.

In order to address the aforementioned drawbacks, Deng et al. (\cite{deng2018learning,deng2020deep}) used the decoder structure of U-Net \cite{ronneberger2015u} to gradually incorporate global information into the shallow features. However, recent research \cite{xuan2022fcl} suggests that semantic information gradually decays as it is fused downward in U-Net structures, diminishing its guiding effect. To simultaneously preserve multi-scale features and generate pixel-level weight matrices, recent works such as FCL \cite{xuan2022fcl} generate a pixel-level weight matrix for each scale feature during multi-scale feature generation. CATS \cite{huan2022cats}, on the other hand, splices multi-scale features and combines spatial and channel information to alleviates edge localization ambiguity. While generating crisp edges with the contribution of tracing loss and a context-aware fusion block (coFusion), it can lead to suppress the nearest neighbor edges. To overcome this limitation, the Double Fusion module is proposed in the current work, which improve efficiently the procedure of coFusion with fewer parameters.


\subsection{Loss Functions}

To improve the performance of edge detection, researchers have proposed various loss functions to optimize the learning process. HED \cite{xie2015hed} is a seminal work in this field that introduced Weighted Cross-Entropy (WCE) as a loss function for the end-to-end supervised learning. However, it is well known that WCE suffers from multiple annotation inconsistencies in the BSDS500 dataset \cite{liu2019RCFext}. In order to address this problem, subsequent studies (e.g.,  \cite{liu2019RCFext,he2022bdcn-ext,su2021pidinet}) propose the WCE+ loss function by ignoring the disputed pixels while measuring WCE.


In recent years, researchers have gradually identified issues with the WCE+ loss function. Due to the significant disparity between the number of edge and non-edge pixels, backpropagation gradients tend to assign larger weights to edge pixels, leading to blurry edges. The problem of imbalanced positive and negative samples is further aggravated by WCE+ where controversial edges are ignored. Therefore, several alternative approaches have been proposed to achieve crisper edge detection by refining the loss function. For instance, \cite{deng2018learning} uses a combination of Dice coefficient and Cross-Entropy. \cite{deng2020deep} goes a step further by incorporating the structural differences between the output and the ground truth using SSIM \cite{zhou2004Image}. Lastly,  \cite{huan2022cats} optimizes the loss function by dividing the image into three categories: edge, confusing, and non-edge pixels.


Although the use of these loss functions has significantly improved edge detection, a crucial issue is ignored: the edge fusion module pays different levels of attention to edge-maps predicted in the preliminary stages. In the current work we address this problem by employing different loss functions to monitor the main architecture and the fusion module (dfuse for TEED) separately, enabling them to capture information at various levels.

\begin{figure*}[t]
  \centering
   \includegraphics[width=0.89\linewidth]{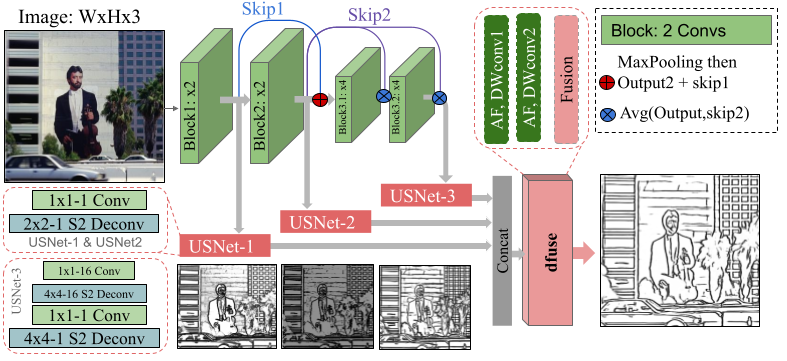}

   \caption{TEED architecture. }
   \label{fig:teed}
\end{figure*}

\section{Proposed Model} \label{sec:teed}

In this section, we present the proposed Tiny and Efficient Edge Detector (TEED) architecture in detail. We begin by introducing the backbone architecture, followed by USNet from DexiNed \cite{soria2023dexined_ext}. Then, we present the Edge fusion module, termed Double Fusion (dfuse), as well as a proposed loss function, named Double Loss (dloss). All components united in TEED provide simplicity, efficiency, and edge detection generalization, making this new approach an effective and efficient edge detection model that reduces training and testing time as well as computational cost.

\subsection{TEED Backbone Architecture}

Since the inception of ResNet \cite{he2016resnet}, Xception \cite{chollet2017xception}, EfficientNet \cite{tan2019efficientnet} architectures powered by dense and skip connections have achieved impressive results improving forward and backward operations from the shallower to the deeper CNN layers. These advantages are plausible in many computer vision tasks. Following these successes, DexiNed \cite{soria2020dexined} and LDC \cite{xsoria2022ldc} use similar architectures for edge detection. The result is a model trained from scratch that still achieves state-of-the-art accuracy. 
The TEED backbone architecture presented in Fig. \ref{fig:teed} is based on LDC \cite{xsoria2022ldc}. It consists of three green blocks (i.e., Block: x2 in Fig. \ref{fig:teed}), each has two standard CNN layers. We did not consider the VGG16 architecture \cite{simonyan2014vgg16} due to its lack of skip-connections, which can reduce the efficacy of edge detection in deeper layers. In TEED we only use $58K$ parameters, which is far lower than DexiNed and LDC, $35M$ and $674K$ parameters respectively. We reached this compact network by reducing the number of convolutional layers and skipping Batch Normalisation (BN). With this reduction also falls the accuracy, to overcome this drawbacks, we consider a new activation function, which is proposed in \cite{wang2022smish}, $\zeta(h)=smish(h)=h\cdot tanh[\ln(1+sigmoid(h))]$, where $h$ is the feature map of the respective layer in TEED. The $smish$ is capable of reducing the lack of efficient training optimization effect without BN. Besides, according to our empirical observations, this non-linearity function improve the RELU function used in the SOTA models.


 Overall, every block of TEED has 2 convolutional layers followed by $\zeta$---$[conv1+smish+conv2+smish]$. The three backbone blocks have $16$, $32$, and $48$ layers, respectively. Then, the output of the first block $h_1$ and the second block $h_2$ are combined by skip connection $skip1$ through the summation operation ($h_1 + h_2$). This skip-connection is applied immediately after the max-pooling operation. Another skip connection, $skip2$, is used to fuse $h_2$ and the sub-outputs of block 3 as follow $(Block3.1+h2)/2$; this is similar to the skip connection in \cite{soria2020dexined} and \cite{xsoria2022ldc}. 
 The outputs of these three blocks feed USNet, as indicated by gray arrows pointing downwards in Fig. \ref{fig:teed}. The convolutional layers used for skip-connections have $1\times 1$ kernel size.


\subsection{USNet}

The USNet module of TEED is similar to DexiNed \cite{soria2020dexined}, with a slight modification: $i)$ we use Xavier initialization in all of USNet layers, $ii)$ we use the same activation function of the backbone ($\zeta$). The outputs of USNet are the edge-map predictions, $\hat{y}_i$, with the same width and height of the input image $x$, where $x\in \mathbb{R}^{W\times H\times 3}$. As illustrated in Fig. \ref{fig:teed}, USNet is composed of 1 convolutional layer (Conv) followed by activation $\zeta$ and 1 deconvolutional layer (Deconv). The kernel size and number of filters can be seen in the bottom left of Fig. \ref{fig:teed}; for instance, "Deconv $2\times2-1$ $S2$" is a deconvolutional layer with a kernel size of $2\times2$, $1$ filter, and the upscale process $\times2$. The USNet-2 has the same architecture as the USNet-1. The size of the output from block 3-2 is $\times4$ down-sampled; hence, USNet-3 has two Conv and 2 Deconv layers as shown in the bottom left of Fig. \ref{fig:teed}. Finally, the edge-map predictions from the respective USNet modules is $\hat{y}_i=\sigma(USNet(h_i))$.


\subsection{Double Fusion}

The edge fusion module in an end-to-end edge detection learning model is typically a CNN or a set of layers that merges edge-maps generated in different scale levels of the backbone network. This module in TEED is named \textbf{dfuse} as illustrated in Fig. \ref{fig:teed}; the predicted edge-map from this module is denoted by $\hat{y}_{dfuse}$. This module is inspired from CATS \cite{huan2022cats}. In terms of efficiency, TEED$+$dfuse comes with just 58K parameters, much less than 99K parameters in TEED$+$coFusion.


The dfuse is composed of two Depth-wise convolutional layers (DWConv), since this approach applies a single convolutional filter to each input channel \cite{guo2019DWconv}, which increments the receptive field but reduce the cost of computation. This module does not use Softmax activation nor group normalization; instead, it employs the $Smish$ activation function \cite{wang2022smish} ($\zeta$) to regularize the weight maps during training. The kernel size of the DWConv is $3\times3-24$, which reduces the spatial dimensions of dfuse. Note that the activation function $\zeta$ is applied before the DWConv layers. The output from the DWConv layers is fused twice:


\begin{center}
\begin{equation}
\begin{aligned}
  h_{dfuse} = \zeta(ewa(DWc_1(\hat{Y}) + DWc_2(\hat{H}_{dfuse}))),\\
  \textrm{where,} \quad \hat{y}_{dfuse} = \sigma(h_{dfuse}),
  \label{eq:doublefusion}
  \end{aligned}
\end{equation}
\end{center}

\noindent and $\sigma$ is the sigmoid function. The output of USNet is denoted as $\hat{Y}=[\hat{y}_1, \hat{y}_3, \hat{y}_3]$ and the feature map of DWConv is $\hat{H}_{dfuse} \in \mathbb{R}^{W\times H \times 24}$. As indicated in eq. (\ref{eq:doublefusion}) two fusions are applied in \textit{dfuse}: 
the first one is the sum of DWConv1 and DWConv2, where the second fusion is applied through element-wise addition ($ewa$) of the first fusion, followed by the activation function $\zeta$. At this stage, $\hat{y}_{dfuse}$ has the same size as the respective $y$.


\subsection{Double Loss}

We introduce an approach to measure the error on the training dataset of pairs $(x, y)$, where $y$ is the ground truth edge map of image $x$. The loss function considered for our end to end training on edge detection is the weighted cross entropy $L_{wce}$, which was originally proposed by HED \cite{xie2015hed} and slightly modified later in BDCN \cite{he2022bdcn-ext}. $L_{wce}$ helps in detecting as many edges as possible. However, if the detected edges are absent in the ground truth, the resulting conflict is reflected in the form of artifacts or noise in the detected edge space. If a model is trained using a ground truth that is visually annotated by humans, $L_{wce}$ may show this drawback; see PiDiNet \cite{su2021pidinet} result in Fig. \ref{fig:banner}.


In order to overcome this problem, CATS \cite{huan2022cats} proposes a new loss function called tracing loss $L_{trcg}$, which is a combination of $L_{wce}$, boundary tracing fusion and texture suppression function. This loss function leads to faster convergence of the model during training, compared to $L_{wce}$, while the predicted edge-maps are clearer and thinner. However, since the ground truth is generated through human visual judgment, some edges may be omitted in the predicted edge-map. As $L_{trcg}$ is relying more on a given ground truth (compared to $L_{wce}$) some true edges may be excluded. In addition, deploying a tiny model such as TEED could result in losing the generalization ability.


To address these drawbacks, TEED employs $L_{wce}$ for comparing the ground-truth $y$ with the outputs of USNets $\{\hat{y_i}\}_1^3$ as well as $L_{trcg}$ for $\hat{y}_{dfuse}$. Therefore, in the one hand, dfuse module is fed with more detected edges thanks to $L_{wce}$, on the other hand $L_{trcg}$ controls the prediction in the dfuse module, $\hat{y}_{dfuse}$. Moreover, the structure of dfuse is developed to reduce the drawbacks of $L_{trcg}$ and leverage the benefits of $L_{wce}$. The global loss (dloss) can be summarized as follow:
\begin{equation}
  L_{dloss} = \sum_{i=1}^3L_{wce}(\hat{y}_i,y) + L_{trcg}(\hat{y}_{dfuse},y).
  \label{eq:Dloss}
\end{equation}


\section{Edge Detection Datasets and Evaluation}
\label{sec:data}

This section presents the datasets used for training and assessing the models discussed in Sec. \ref{sec:expe} together with a brief overview of the metrics employed for evaluation.



\subsection{Datasets for Training TEED}

Due to the impressive results of the recently released dataset for edge detection, Barcelona Images for Perceptual Edge Detection (BIPEDv2) \cite{soria2020dexined,soria2023dexined_ext}  referred to as BIPED in this manuscript, we trained all models with this dataset.
\textbf{BIPED} was firstly introduced in \cite{soria2020dexined}. It contains 250 images in high definition ($720\times1280$): 50 images of which were selected by the authors for testing and the rest for training and validation. 
The data augmentation procedure used in LDC \cite{xsoria2022ldc} is implemented for TEED.


\subsection{Dataset for Testing Edge Generalization}

The well known datasets BSDS500 \cite{arbelaez2011bsds500}, PASCAL-Context \cite{mottaghi2014PASCALcontext}, and NYUD \cite{Gupta_2013NYUDv2} are usually considered for training and evaluating edge detection methods. However, based on our understanding and analysis of the literature \cite{li2019noBSDS, Hou_2013NObsds, soria2023dexined_ext}, this assumption is wrong, as these datasets are not intended for edge detection. The ground truths in these datasets are prepared for boundary detection and/or image segmentation \cite{li2019edgeSupport}. Only BIPED \cite{soria2023dexined_ext} and BRIND \cite{pu2021brind} have been proposed to tackle the edge detection problem.




Hence, in order to test whether a trained model can generalize by detecting edges on images from different scenes, a new dataset is proposed: Unified Dataset for Edge Detection (UDED). This UDED is created with the purpose of reducing the evaluation procedure by selecting images from datasets focused on low- and mid-level tasks, and carefully annotating perceptual edges. The proposed UDED contains 30 images selected from: BIPED \cite{soria2020dexined}, BSDS500 \cite{arbelaez2011bsds500}, BSDS300 \cite{martin2004mBSDS300ext}, DIV2K \cite{Ignatov2018DIV2K}, WIREFRAME \cite{Kun2018wireframe}, CID \cite{grigorescu2003cid}, CITYSCAPES \cite{Cordts2016Cityscapes}, ADE20K \cite{zhou2019ade20k}, MDBD \cite{mely2016mdbd}, NYUD \cite{Gupta_2013NYUDv2}, THANGKA \cite{Yanchun2021Thangka}, PASCAL-Context \cite{mottaghi2014PASCALcontext}, SET14 and URBAN100 \cite{huang2015SRdataset}, and the cameraman image. The image selection process consists on computing the Inter-Quartile Range (IQR) intensity value on all the images, images larger than $720\times720$ pixels were not considered. Then, the images are sorted by this IQR value and 30 of them uniformly selected from that sorted list---around 2 images per dataset have been selected.

Finally, since edge detection is a low-level process, required for other tasks (e.g., image super-resolution guidance \cite{gupta2020edgeTherSR} and sketch-based image retrieval \cite{yu2016EdgeSketch}), we propose to conduct an application-oriented evaluation that validates the generalization of UDED.


\subsection{Metrics for Quantitative Evaluation}

In the current work, we focus on the most commonly used metrics for quantitatively evaluating edge detection methods: Optimal Dataset Scale (ODS) and Optimal Image Scale (OIS) \cite{martin2004mBSDS300ext}. Additionally, Peak Signal to Noise Ratio (PSNR), Mean Square Error (MSE), and Mean Absolute Error (MAE) are also considered for the quantitative comparison as they have recently been suggested in \cite{sadykova2017EdgeMetric, eser2019edgeMetric, tariq2021edgeMetric, jing2022edgeMetric}. 

\section{Experiments} \label{sec:expe}

In general, an edge detector based on CNN is validated through training and testing in different dataset (e.g., BSDS \cite{arbelaez2011bsds500}, NYUDv2 \cite{Gupta_2013NYUDv2}). Our manuscript proposes a new methodology for the edge detection model evaluation. We suggest evaluating it only in a dataset especially prepared for edge detection. This dataset should be designed in a way that is efficient and gives results in a short period of time, which is not possible in the test set of the state of the art, nowadays. Therefore, we propose evaluating on the UDED dataset, presented in Sec. \ref{sec:data}. To validate the effectiveness of UDED, a downstream task for sketch image retrieval is considered \cite{yu2016EdgeSketch, song2017edgeSketch2}. Implementation details are given below, followed by an ablation study. Later, we present both quantitative and qualitative results using the UDED dataset. Finally, TEED is validated in sketch image retrieval.

\subsection{Implementation Details}

TEED is implemented in PyTorch and trained on an NVIDIA 3090 GPU. TEED training is based on Adam optimizer \cite{bengio2015adam}, a batch size of 8, an initial learning rate of $8e-4$ changed to $8e-5$ at epoch 5, a weight decay of $2e-4$. Validation results at epoch $6$ are reported. Despite the fact that BIPED provides binary annotations, due to the interpolations when edge maps are built floating point values appear; hence we apply a transformation to edge annotations $y$ by adding $0.2$ to values greater than $0.1$, and then clipping the results to [0,1], similar to LDC \cite{xsoria2022ldc}. A Lenovo Yoga C740-15IML laptop with an Intel i5-10210U processor is used to report the FPS values.



\subsection{Ablation Study} \label{sub:ablation}

\begin{table*}
\setlength\tabcolsep{2pt} 
\centering
\begin{tabular}{|lcccccccccccccc|}
\hline
B2& B3&\#P&Train-data&$L_{wce}$&$L_{trcg}$&coF&DF&Conv&DWConv&Relu&Tanh&Smish&ODS&OIS\\
 \hline \hline
&\checkmark&$99K$&BIPED&&\checkmark&\checkmark&&\checkmark&&&&\checkmark&.810&.842\\
 &\checkmark&$99K$&BIPED&\checkmark&\checkmark&\checkmark&&\checkmark&&&&\checkmark&.821&\textbf{.854}\\
  &\checkmark&\cblue{58K}&BIPED&\checkmark&\checkmark&&Hap&&\checkmark&&&\checkmark&.814&.84\\
  &\checkmark&$60K$&BIPED&\checkmark&\checkmark&&EWA&\checkmark&&&&\checkmark&\cblue{.825}&\cblue{.851}\\
 \hline
&\checkmark&\cblue{58K}&BIPED&\checkmark&\checkmark&&EWA&&\checkmark&\checkmark&&&.816&.827\\
 &\checkmark&\cblue{58K}&BIPED&\checkmark&\checkmark&&EWA&&\checkmark&&\checkmark&&.815&.837\\
\checkmark&&\textbf{17K}&BIPED&\checkmark&\checkmark&&EWA&&\checkmark&&&\checkmark&.796&.823\\

&\checkmark&\cblue{58K}&BIPED&\checkmark&\checkmark&&EWA&&\checkmark&&&\checkmark&\textbf{.828}&.842\\




\hline

\hline
\end{tabular}
\caption{Detailed ablation study of TEED backbone, Double-Loss and Double-Fusion. }
\label{tab:abla}
\end{table*}

In this section various components of the TEED model are analyzed, using BIPED \cite{soria2023dexined_ext} as training data. Table \ref{tab:abla} shows the different configurations of TEED, starting in the first column with $B2$, as showing in Fig. \ref{fig:teed}, TEED is composed of three blocks, and $B2$ represents just 2 blocks of the proposal and $B3$ corresponds to the results when all blocks are considered. The table presents information about the number of parameters ($\#P$), loss functions used in training ($L_{trcg}$ proposed for CAST \cite{huan2022cats} and $L_{wce}$ the weighted cross entropy loss), fusion modules ($coF$), activation functions (including $Relu$, $Tanh$, and $Smish$). Column $Conv$ corresponds to the standard convolution layer used in the $dfuse$, and the column $DWConv$ corresponds to the Depth Wise Convolution used on the fusion module. The last two columns correspond to the standard metrics used for the edge detection quantitative evaluation.


Table \ref{tab:abla} is divided into two sections. The top section shows results of TEED using the first configuration. For instance, the first row in the top section shows the results of TEED using only the loss function from CATS with the same fusion module (coFusion), and the Smish activation function in the TEED backbone. The resulting ODS is $0.810$, and this version of TEED has 99K parameters. Starting from the third row in the top section of Table \ref{tab:abla}, we begin using the dfuse module, which reduces the number of parameters by ~40K; by using DWConv, we reduce the number of parameters from 60K to 58K for TEED. Overall, we can see that using dloss, dfuse, and the Smish activation function in the TEED model contribute to both efficiency and accuracy. 


\subsection{Quantitative Results}

Based on the quantitative results from DexiNed \cite{soria2023dexined_ext} and LDC \cite{xsoria2022ldc}, we trained all models with BIPED. Table \ref{tab:uded} presents results from TEED and the state-of-the-art models with less than \textbf{1M parameters}; all these models are trained with BIPED and evaluated with UDED. The approaches considered in the comparison are as follow: the block 2 of BDCN \cite{he2022bdcn-ext}---BDCN-B2; three versions of PiDiNet \cite{su2021pidinet}---the standard PiDiNet, PidiNed-Small, and PiDiNet-tiny-L; TIN \cite{wibisono2020tin}; LDC \cite{xsoria2022ldc}; Canny edge detector \cite{canny1986canny} and TEEDup---we up-scale the input image ($x$), to $1.5$ before feeding the model. Results from DexiNed edge detection model \cite{soria2023dexined_ext} are provided just for reference. This table shows ODS and OIS, the last epoch used for evaluation, number of parameters (\#P), time of training till reaching the last epoch (Train-time), Frames Per Second (FPS), Mean Square Error (MSE), Mean Absolute Error (MAE), and Peak-Signal-to-Noise-Ratio (PSNR). The results in ODS and OIS are from edge-maps after applying NMS. Results from MSE, MAE, and PSNR are before applying NMS, this process also let us know which edge-map has less artifacts or noises.

As shown in the table, TEED and TEEDup achieve the best results in all evaluation criteria with only 6 epochs and a training time of less than 30 minutes. In contrast, all other approaches require between 10 and 53 hours of training to achieve similar performance. It should be noted that TEED is the architecture with less number of parameters. The second smaller architecture (PiDiNet-tiny-L \cite{su2021pidinet}) requires 30 hours of training, while TEED needs only \textbf{30 minutes}.


\begin{table*}
\begin{center}
\begin{tabular}{|l|cccc|cc|ccc|}
\hline
Method & $\downarrow$Epoch& $\downarrow$\#P  & $\downarrow$Train-time & $\uparrow$FPS& $\uparrow$ODS & $\uparrow$OIS& $\downarrow$MSE & $\downarrow$MAE &$\uparrow$PSNR \\
\hline\hline
Canny \cite{canny1986canny} & --&--&--&392.5&.742&.743&---&---&---\\ 
\hline
DexiNed \cite{soria2023dexined_ext}& \textit{11}	&\textit{35M}	&$\sim$\textit{20 hours	}	&\textit{.34}	&\textit{.815}	&\textit{.826}	&\textit{.095}	&\textit{.149}	&\textit{10.799} \\
PiDiNet \cite{su2021pidinet}&20	&710K	&$\sim$53 hours		&.67	&.812	&.824	&.126	&.194	&9.49 \\
LDC \cite{xsoria2022ldc} & \cblue{16}	&674K	&$\sim$\cblue{10 hours}	&\cblue{2.57}	&.817	&.838 &.084	&\cblue{.134}&11.268\\
BDCN-B2 \cite{he2022bdcn-ext}&20	&268K	&--		&1.97	&.821	&.839	&.136	&.205	&9.229\\
TIN \cite{wibisono2020tin}&1.6M	&244K	&$\sim$14 Hours	&1.2	&.803	&.827	&.094	&.16	&10.734\\
PiDiNet-small \cite{su2021pidinet}&20	&184K	&$\sim$40 hours	&---	&.821	&.834	&.133	&.202	&9.174\\
PiDiNet-tiny-L \cite{su2021pidinet}&20	&\cblue{73K}	&$\sim$30 hours&1.59	&.821	&.834	&.136	&.21	&9.182\\
\hline
TEED (Ours)&\textbf{6}	&\textbf{58K}	&$\sim$\textbf{30 min}&\textbf{2.6}	&\cblue{.828}&\cblue{.842}&\cblue{.073}	&\textbf{.107}	&\cblue{11.965}\\
TEEDup (Ours)&\textbf{6}	&\textbf{58K}	&$\sim$\textbf{30 min}&1.94	&\textbf{.834}	&\textbf{.847}	&\textbf{.071}	&\textbf{.107}	&\textbf{12.05}\\
\hline
\end{tabular}
\end{center}
\caption{Results when models are trained with BIPED \cite{soria2023dexined_ext} but evaluated with the proposed UDED dataset.}
\label{tab:uded}
\end{table*}


\subsection{Qualitative Results}

\begin{figure*}[t]
  \centering

   \includegraphics[width=0.96\textwidth, height=0.92\textheight]{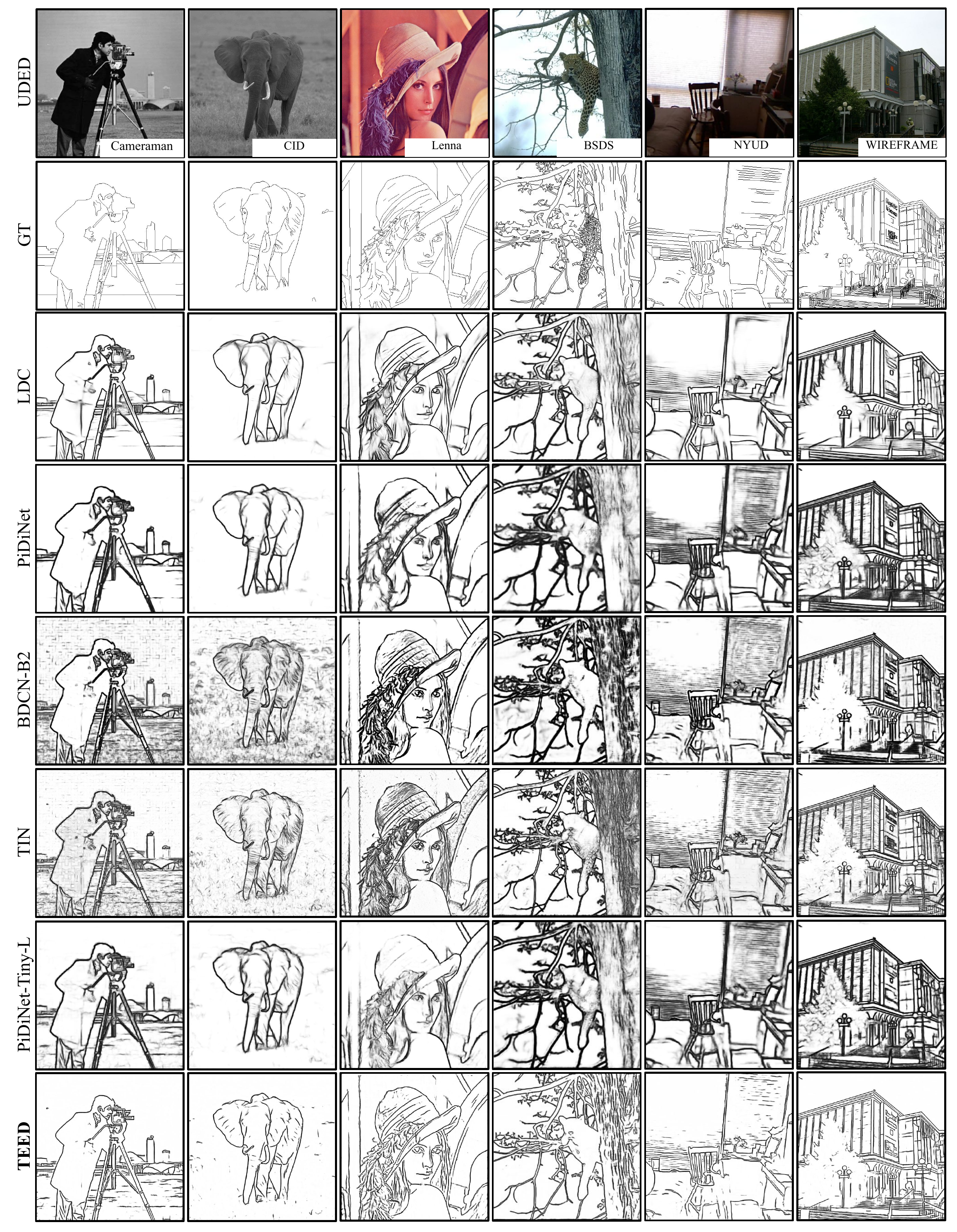}

   \caption{Edge-maps predicted with the lightweight SOTA models and TEED---images from UDED dataset.}
   \label{fig:qualres}
\end{figure*}

Figure \ref{fig:qualres} presents six images from UDED dataset for the perceptual judgement. The images used in UDED come from different datasets and the selection procedure is detailed in Sec \ref{sec:data}. We can see that from the third column, the images used in the comparison are challenging; for instance, the image from the last column has a large number of edges annotated in the ground truth, and most of the models used for comparison predict edge-maps with noise. In conclusion, it can be said that the compact TEED architecture excels at detecting as many edges as possible. TEED is able to obtain superior edge-maps compared to state-of-the-art approaches in all the images shown in Fig. \ref{fig:qualres}; actually, not only more edges are detected but also \textbf{thinner} and \textbf{cleaner} edge-maps are obtained.

\subsection{Discussion}\label{sub:diss}

\begin{table}
\scriptsize 
\centering
\begin{tabular}{ |cccccc| }    
\hline
&& \multicolumn{2}{c}{QMUL-Shoe \cite{yu2016EdgeSketch}}& \multicolumn{2}{c|}{QMUL-Chair \cite{yu2016EdgeSketch}}\\
Method &\#P& Top1 & Top10& Top1 & Top10 \\
\hline\hline
TripletSN \cite{yu2016EdgeSketch}&---&.3913	&.8783&.6907	&.9794\\
BDCN-BSDS &16.3M	&.3913	&.8348&.6391	&.9896\\
BDCN-BIPED&16.3M	&.513	&.8869&.8144	&.9896\\
PiDiNet-BIPED&710K&.5043 &.8347 &.8041	&\textbf{1}\\
TEED-BIPED&\textbf{58K}&\textbf{.5217}	&\textbf{.8957}&\textbf{.835}	&\textbf{1}\\
\hline
\end{tabular}
\caption{Sketch image retrieval results in QMUL Shoe and Chair datasets \cite{yu2016EdgeSketch}. BDCN-BSDS is the model trained with BSDS \cite{Hou_2013NObsds}, BDCN-BIPED stands for the model trained with BIPED \cite{soria2023dexined_ext}.}
\label{tab:disc1}
\end{table}

\begin{table}
\scriptsize 
\centering
\begin{tabular}{ |cccccc| }    
\hline
&& \multicolumn{2}{c}{QMUL-ShoeV2 \cite{song2017edgeSketch2}}& \multicolumn{2}{c|}{QMUL-ChairV2 \cite{song2017edgeSketch2}}\\
Method &\#P& Top1 & Top10& Top1 & Top10 \\
\hline\hline
HOLEF \cite{song2017edgeSketch2}& ---&\textbf{.6174}&.9478 &.8144	&.9588\\
BDCN-BSDS&16.3M	&.3826	&.8869	&.6804	&.9896\\
BDCN-BIPED&16.3M	&.5565	&.8956&\textbf{.8969}	&.9896\\
PiDiNet-BIPED&710K &.5304	&.8869 &.8762	&\textbf{1}\\
TEED-BIPED&\textbf{58K}&.5565	&\textbf{.9565}&.8865	&\textbf{1}\\
\hline
\end{tabular}
\caption{Sketch image retrieval results in the updated QMUL shoe and chair datasets reported in  HOLEF \cite{song2017edgeSketch2}.}
\label{tab:disc2}
\end{table}

Since the 1980s, the edge detection evaluation on a set of images with corresponding ground truths, has been challenging \cite{ziou1998edgeOverview1, yang2022edgeOverview}. This is because edge maps are not the ultimate goal, but rather they are used for higher level tasks \cite{soria2023dexined_ext} of computer vision and image processing. For addressing this issue, we propose the UDED dataset, which includes a small yet diverse set of images with different intensities to evaluate the performance under various scenarios.


In order to validate our dataset, we consider a subsequent use of the edge-maps generated from TEED. Testing predicted edge-maps in some application is an effective way to validate an edge detector. Hence, we consider sketch based image retrieval \cite{yu2016EdgeSketch, song2017edgeSketch2} to compare the edge-map contribution in the first and last versions of QMUL-chair and QMUL-shoe datasets; results are depicted in Table \ref{tab:disc1} and \ref{tab:disc2}. These results show that reaching the best performance in the most widely used datasets does not guarantee its effectiveness in the subsequent tasks. TEED achieves the best results in sketch image retrieval task using only 58K parameters, which is less than 9\% of PiDNet's \cite{su2021pidinet} parameters and less than 0.3\% of BDCN's parameters. The results suggest that TEED has a strong generalization capability, as it performs well on new datasets like QMUL-chair and QMUL-shoe, even though it was only trained on BIPED.


\section{Conclusions} \label{sec:conc}

This paper presents TEED, a deep learning-based edge detector that produces edge-maps that mimics human visual perception. The performance of TEED is evaluated using various metrics and a subsequent task, sketch-based image retrieval, and compared with SOTA edge detectors. Our results demonstrate that TEED outperforms other edge detectors in terms of accuracy, while requiring significantly fewer parameters and being easy to train within few epochs. Additionally, our experiments show that the proposed UDED dataset is a useful tool for validating edge detectors across different scenarios. Overall, the results suggest that TEED is a highly effective and efficient edge detector that could be used in a wide range of computer vision applications.

\section{Acknowledgements}
\noindent This material is based upon work supported by the Air Force Office of Scientific Research under award number FA9550-22-1-0261; and partially supported by the Grant PID2021-128945NB-I00 funded by MCIN/AEI/10.13039/501100011033 and by ``ERDF A way of making Europe"; the ``CERCA Programme / Generalitat de Catalunya"; and the ESPOL project CIDIS-12-2022.

{\small
\bibliographystyle{ieee_fullname}
\bibliography{egbib}

\begin{thebibliography}{10}\itemsep=-1pt

\bibitem{arbelaez2011bsds500}
Pablo Arbelaez, Michael Maire, Charless Fowlkes, and Jitendra Malik.
\newblock Contour detection and hierarchical image segmentation.
\newblock {\em Transactions on Pattern Analysis and Machine Intelligence},
  33(5):898--916, 2011.

\bibitem{Basu2002edge_survey}
M. {Basu}.
\newblock Gaussian-based edge-detection methods-a survey.
\newblock {\em Transactions on Systems, Man, and Cybernetics, Part C
  (Applications and Reviews)}, 32(3), 2002.

\bibitem{canny1986canny}
John Canny.
\newblock A computational approach to edge detection.
\newblock {\em IEEE Transactions on pattern analysis and machine intelligence},
  (6):679--698, 1986.

\bibitem{chollet2017xception}
Fran{\c{c}}ois Chollet.
\newblock Xception: Deep learning with depthwise separable convolutions.
\newblock In {\em Proceedings of the IEEE conference on computer vision and
  pattern recognition}, pages 1251--1258, 2017.

\bibitem{Cordts2016Cityscapes}
Marius Cordts, Mohamed Omran, Sebastian Ramos, Timo Rehfeld, Markus Enzweiler,
  Rodrigo Benenson, Uwe Franke, Stefan Roth, and Bernt Schiele.
\newblock The cityscapes dataset for semantic urban scene understanding.
\newblock In {\em Proc. of the IEEE Conference on Computer Vision and Pattern
  Recognition (CVPR)}, 2016.

\bibitem{deng2020deep}
Ruoxi Deng and Shengjun Liu.
\newblock Deep structural contour detection.
\newblock In {\em Proceedings of the 28th ACM international conference on
  multimedia}, pages 304--312, 2020.

\bibitem{deng2018learning}
Ruoxi Deng, Chunhua Shen, Shengjun Liu, Huibing Wang, and Xinru Liu.
\newblock Learning to predict crisp boundaries.
\newblock In {\em Proceedings of the European Conference on Computer Vision
  (ECCV)}, pages 562--578, 2018.

\bibitem{du2022EdgeMedSeg}
Xiaogang Du, Yinyin Nie, Fuhai Wang, Tao Lei, Song Wang, and Xuejun Zhang.
\newblock Al-net: Asymmetric lightweight network for medical image
  segmentation.
\newblock {\em Frontiers in Signal Processing}, 2, 2022.

\bibitem{eser2019edgeMetric}
SERT Eser and AVCI Derya.
\newblock A new edge detection approach via neutrosophy based on maximum norm
  entropy.
\newblock {\em Expert Systems with Applications}, 115:499--511, 2019.

\bibitem{grigorescu2003cid}
Cosmin Grigorescu, Nicolai Petkov, and Michel~A Westenberg.
\newblock Contour detection based on nonclassical receptive field inhibition.
\newblock {\em Transactions on Image Processing}, 12(7):729--739, 2003.

\bibitem{guo2019DWconv}
Yunhui Guo, Yandong Li, Liqiang Wang, and Tajana Rosing.
\newblock Depthwise convolution is all you need for learning multiple visual
  domains.
\newblock In {\em Proceedings of the AAAI Conference on Artificial
  Intelligence}, volume~33, pages 8368--8375, 2019.

\bibitem{gupta2020edgeTherSR}
Honey Gupta and Kaushik Mitra.
\newblock Pyramidal edge-maps and attention based guided thermal
  super-resolution.
\newblock In {\em Computer Vision--ECCV 2020 Workshops: Glasgow, UK, August
  23--28, 2020, Proceedings, Part III 16}, pages 698--715. Springer, 2020.

\bibitem{Gupta_2013NYUDv2}
Saurabh Gupta, Pablo Arbelaez, and Jitendra Malik.
\newblock Perceptual organization and recognition of indoor scenes from rgb-d
  images.
\newblock In {\em Conference on Computer Vision and Pattern Recognition}, June
  2013.

\bibitem{he2022bdcn-ext}
Jianzhong He, Shiliang Zhang, Ming Yang, Yanhu Shan, and Tiejun Huang.
\newblock Bdcn: Bi-directional cascade network for perceptual edge detection.
\newblock {\em IEEE Transactions on Pattern Analysis and Machine Intelligence},
  44(1):100--113, 2022.

\bibitem{he2016resnet}
Kaiming He, Xiangyu Zhang, Shaoqing Ren, and Jian Sun.
\newblock Deep residual learning for image recognition.
\newblock In {\em Proceedings of the IEEE conference on computer vision and
  pattern recognition}, pages 770--778, 2016.

\bibitem{Hou_2013NObsds}
Xiaodi Hou, Alan Yuille, and Christof Koch.
\newblock Boundary detection benchmarking: Beyond f-measures.
\newblock In {\em Conference on Computer Vision and Pattern Recognition}, June
  2013.

\bibitem{huan2022cats}
Linxi Huan, Nan Xue, Xianwei Zheng, Wei He, Jianya Gong, and Gui-Song Xia.
\newblock Unmixing convolutional features for crisp edge detection.
\newblock {\em IEEE Transactions on Pattern Analysis and Machine Intelligence},
  44(10):6602--6609, 2022.

\bibitem{huang2015SRdataset}
Jia-Bin Huang, Abhishek Singh, and Narendra Ahuja.
\newblock Single image super-resolution from transformed self-exemplars.
\newblock In {\em Proceedings of the IEEE conference on computer vision and
  pattern recognition}, pages 5197--5206, 2015.

\bibitem{Kun2018wireframe}
Kun Huang, Yifan Wang, Zihan Zhou, Tianjiao Ding, Shenghua Gao, and Yi Ma.
\newblock Learning to parse wireframes in images of man-made environments.
\newblock In {\em CVPR}, June 2018.

\bibitem{Ignatov2018DIV2K}
Andrey Ignatov, Radu Timofte, et~al.
\newblock Pirm challenge on perceptual image enhancement on smartphones:
  report.
\newblock In {\em European Conference on Computer Vision (ECCV) Workshops},
  January 2019.

\bibitem{jing2022edgeMetric}
Junfeng Jing, Shenjuan Liu, Gang Wang, Weichuan Zhang, and Changming Sun.
\newblock Recent advances on image edge detection: A comprehensive review.
\newblock {\em Neurocomputing}, 2022.

\bibitem{khan2022ViTsurvey}
Salman Khan, Muzammal Naseer, Munawar Hayat, Syed~Waqas Zamir, Fahad~Shahbaz
  Khan, and Mubarak Shah.
\newblock Transformers in vision: A survey.
\newblock {\em ACM computing surveys (CSUR)}, 54(10s):1--41, 2022.

\bibitem{bengio2015adam}
Diederik~P. Kingma and Jimmy Ba.
\newblock Adam: {A} method for stochastic optimization.
\newblock In Yoshua Bengio and Yann LeCun, editors, {\em 3rd International
  Conference on Learning Representations, {ICLR} 2015, San Diego, CA, USA, May
  7-9, 2015, Conference Track Proceedings}, 2015.

\bibitem{li2019noBSDS}
Ou Li and Peng-Lang Shui.
\newblock Noise-robust color edge detection using anisotropic morphological
  directional derivative matrix.
\newblock {\em Signal Processing}, 165:90--103, 2019.

\bibitem{li2019edgeSupport}
Ou Li and Peng-Lang Shui.
\newblock Noise-robust color edge detection using anisotropic morphological
  directional derivative matrix.
\newblock {\em Signal Processing}, 165:90--103, 2019.

\bibitem{liu2019RCFext}
Y. {Liu}, M. {Cheng}, X. {Hu}, J. {Bian}, L. {Zhang}, X. {Bai}, and J. {Tang}.
\newblock Richer convolutional features for edge detection.
\newblock {\em IEEE Transactions on Pattern Analysis and Machine Intelligence},
  41(8):1939--1946, 2019.

\bibitem{Yanchun2021Thangka}
Yanchun Ma, Yongjian Liu, Qing Xie, Shengwu Xiong, Lihua Bai, and Anshu Hu.
\newblock A tibetan thangka data set and relative tasks.
\newblock {\em Image and Vision Computing}, 108:104125, 2021.

\bibitem{maini2009edgeOverview3}
Raman Maini and Himanshu Aggarwal.
\newblock Study and comparison of various image edge detection techniques.
\newblock {\em International Journal of Image Processing}, 3(1), 2009.

\bibitem{martin2004mBSDS300ext}
D.R. Martin, C.C. Fowlkes, and J. Malik.
\newblock Learning to detect natural image boundaries using local brightness,
  color, and texture cues.
\newblock {\em Transactions on Pattern Analysis and Machine Intelligence},
  26(5):530--549, 2004.

\bibitem{mely2016mdbd}
David~A M{\'e}ly, Junkyung Kim, Mason McGill, Yuliang Guo, and Thomas Serre.
\newblock A systematic comparison between visual cues for boundary detection.
\newblock {\em Vision Research}, 120, 2016.

\bibitem{mottaghi2014PASCALcontext}
Roozbeh Mottaghi, Xianjie Chen, Xiaobai Liu, Nam-Gyu Cho, Seong-Whan Lee, Sanja
  Fidler, Raquel Urtasun, and Alan Yuille.
\newblock The role of context for object detection and semantic segmentation in
  the wild.
\newblock In {\em Conference on Computer Vision and Pattern Recognition}, 2014.

\bibitem{pu2021brind}
Mengyang Pu, Yaping Huang, Qingji Guan, and Haibin Ling.
\newblock Rindnet: Edge detection for discontinuity in reflectance,
  illumination, normal and depth.
\newblock In {\em International Conference on Computer Vision}, pages
  6879--6888, 2021.

\bibitem{Pu2022edter}
Mengyang Pu, Yaping Huang, Yuming Liu, Qingji Guan, and Haibin Ling.
\newblock Edter: Edge detection with transformer.
\newblock In {\em Proceedings of the IEEE/CVF Conference on Computer Vision and
  Pattern Recognition (CVPR)}, pages 1402--1412, June 2022.

\bibitem{ronneberger2015u}
Olaf Ronneberger, Philipp Fischer, and Thomas Brox.
\newblock U-net: Convolutional networks for biomedical image segmentation.
\newblock In {\em International Conference on Medical image computing and
  computer-assisted intervention}, pages 234--241. Springer, 2015.

\bibitem{sadykova2017EdgeMetric}
Diana Sadykova and Alex~Pappachen James.
\newblock Quality assessment metrics for edge detection and edge-aware
  filtering: A tutorial review.
\newblock In {\em 2017 International Conference on Advances in Computing,
  Communications and Informatics (ICACCI)}, pages 2366--2369. IEEE, 2017.

\bibitem{simonyan2014vgg16}
Karen Simonyan and Andrew Zisserman.
\newblock Very deep convolutional networks for large-scale image recognition.
\newblock {\em arXiv preprint arXiv:1409.1556}, 2014.

\bibitem{sobel1972sobelmethod}
Irwin Sobel.
\newblock Camera models and machine perception.
\newblock Technical report, Computer Science Department, Technion, 1972.

\bibitem{song2017edgeSketch2}
Jifei Song, Qian Yu, Yi-Zhe Song, Tao Xiang, and Timothy~M Hospedales.
\newblock Deep spatial-semantic attention for fine-grained sketch-based image
  retrieval.
\newblock In {\em Proceedings of the IEEE international conference on computer
  vision}, pages 5551--5560, 2017.

\bibitem{xsoria2022ldc}
Xavier Soria, Gonzalo Pomboza-Junez, and Angel~Domingo Sappa.
\newblock Ldc: Lightweight dense cnn for edge detection.
\newblock {\em IEEE Access}, 10:68281--68290, 2022.

\bibitem{soria2020dexined}
Xavier Soria, Edgar Riba, and Angel Sappa.
\newblock Dense extreme inception network: Towards a robust cnn model for edge
  detection.
\newblock In {\em Winter Conference on Applications of Computer Vision}, 2020.

\bibitem{soria2023dexined_ext}
Xavier Soria, Angel Sappa, Patricio Humanante, and Arash Akbarinia.
\newblock Dense extreme inception network for edge detection.
\newblock {\em Pattern Recognition}, 139:109461, 2023.

\bibitem{su2021pidinet}
Zhuo Su, Wenzhe Liu, Zitong Yu, Dewen Hu, Qing Liao, Qi Tian, Matti
  Pietikäinen, and Li Liu.
\newblock Pixel difference networks for efficient edge detection.
\newblock In {\em International Conference on Computer Vision (ICCV)}, pages
  5097--5107, 2021.

\bibitem{tan2019efficientnet}
Mingxing Tan and Quoc Le.
\newblock Efficientnet: Rethinking model scaling for convolutional neural
  networks.
\newblock In {\em International conference on machine learning}, pages
  6105--6114. PMLR, 2019.

\bibitem{tariq2021edgeMetric}
Nazish Tariq, Rostam~Affendi Hamzah, Theam~Foo Ng, Shir~Li Wang, and Haidi
  Ibrahim.
\newblock Quality assessment methods to evaluate the performance of edge
  detection algorithms for digital image: A systematic literature review.
\newblock {\em IEEE Access}, 9:87763--87776, 2021.

\bibitem{wang2022smish}
Xueliang Wang, Honge Ren, and Achuan Wang.
\newblock Smish: A novel activation function for deep learning methods.
\newblock {\em Electronics}, 11(4):540, 2022.

\bibitem{zhou2004Image}
Zhou Wang, A.C. Bovik, H.R. Sheikh, and E.P. Simoncelli.
\newblock Image quality assessment: from error visibility to structural
  similarity.
\newblock {\em IEEE Transactions on Image Processing}, 13(4):600--612, 2004.

\bibitem{wibisono2020tin}
Jan~Kristanto Wibisono and Hsueh-Ming Hang.
\newblock Traditional method inspired deep neural network for edge detection.
\newblock In {\em IEEE International Conference on Image Processing (ICIP)},
  pages 678--682, 2020.

\bibitem{xie2015hed}
Saining Xie and Zhuowen Tu.
\newblock Holistically-nested edge detection.
\newblock In {\em Proceedings of the IEEE international conference on computer
  vision}, pages 1395--1403, 2015.

\bibitem{xuan2022fcl}
Wenjie Xuan, Shaoli Huang, Juhua Liu, and Bo Du.
\newblock Fcl-net: Towards accurate edge detection via fine-scale corrective
  learning.
\newblock {\em Neural Networks}, 145:248--259, 2022.

\bibitem{yang2022edgeOverview}
Daipeng Yang, Bo Peng, Zaid Al-Huda, Asad Malik, and Donghai Zhai.
\newblock An overview of edge and object contour detection.
\newblock {\em Neurocomputing}, 2022.

\bibitem{yang2019SRreview}
Wenming Yang, Xuechen Zhang, Yapeng Tian, Wei Wang, Jing-Hao Xue, and Qingmin
  Liao.
\newblock Deep learning for single image super-resolution: A brief review.
\newblock {\em IEEE Transactions on Multimedia}, 21(12):3106--3121, 2019.

\bibitem{yu2016EdgeSketch}
Qian Yu, Feng Liu, Yi-Zhe Song, Tao Xiang, Timothy~M Hospedales, and
  Chen-Change Loy.
\newblock Sketch me that shoe.
\newblock In {\em Proceedings of the IEEE Conference on Computer Vision and
  Pattern Recognition}, pages 799--807, 2016.

\bibitem{zhai2022scalingVT}
Xiaohua Zhai, Alexander Kolesnikov, Neil Houlsby, and Lucas Beyer.
\newblock Scaling vision transformers.
\newblock In {\em CVPR}, pages 12104--12113, 2022.

\bibitem{zhou2019ade20k}
Bolei Zhou, Hang Zhao, Xavier Puig, Tete Xiao, Sanja Fidler, Adela Barriuso,
  and Antonio Torralba.
\newblock Semantic understanding of scenes through the ade20k dataset.
\newblock {\em International Journal of Computer Vision}, 127:302--321, 2019.

\bibitem{ziou1998edgeOverview1}
Djemel Ziou, Salvatore Tabbone, et~al.
\newblock Edge detection techniques-an overview.
\newblock {\em Pattern Recognition and Image Analysis C/C of Raspoznavaniye
  Obrazov I Analiz Izobrazhenii}, 8, 1998.

\end{thebibliography}
}

\end{document}